# Disease Classification within Dermascopic Images Using features extracted by ResNet50 and classification through DeepForest

Suhita Ray
Indian Institute of Technology, Kharagpur
Assam, India
suhita.ray@gmail.com

*Abstract* – In this report we propose a classification technique for skin lesion images as a part of our submission for *ISIC 2018 Challenge in Skin Lesion Analysis Towards Melanoma Detection.* Our data was extracted from the "ISIC 2018: Skin Lesion Analysis Towards Melanoma Detection" grand challenge datasets [1][2]. The features are extracted through a Convolutional Neural Network, in our case ResNet50[3] and then using these features we train a DeepForest, having cascading layers, to classify our skin lesion images. We know that Convolutional Neural Networks are a state-of-the-art technique in representation learning for images, with the convolutional filters learning to detect features from images through backpropagation. These features are then usually fed to a classifier like a softmax layer or other such classifiers for classification tasks. In our case we do not use the traditional backpropagation method and train a softmax layer for classification. Instead, we use Deep Forest[4], a novel decision tree ensemble approach with performance highly competitive to deep neural networks in a broad range of tasks. Thus we use a ResNet50 to extract the features from skin lesion images and then use the Deep Forest to classify these images. This method has been used because Deep Forest has been found to be hugely efficient in areas where there are only small-scale training data available. Also as the Deep Forest network decides its complexity by itself, it also caters to the problem of dataset imbalance we faced in this problem.

## 1 INTRODUCTION

Convolutional Neural Networks are one of the most popular technique used in representation learning. Since, it is very difficult for a human to accurately identify the features in an image which differ from one image to another in a multiclass classification problem, Convolutional neural networks have gained a lot of popularity in image classification tasks. But, one of the primary limitations of Convolutional neural network is that a large amount of labelled data must be available for Convolutional neural networks to perform satisfactorily. Another limitation is that performance of convolutional neural networks increase with model complexity, which automatically lead to the requirement of powerful computational facilities for training process. However, as proposed by Zhi-Hua Zhou *et al.* 2018[4], Deep Forest is a novel decision tree ensemble approach, where Deep Forest, henceforth refered to as gcForest, can work well even where there are only small-scale training data. The number of cascade levels can be adaptively determined such that the model complexity can be automatically set, enabling gcForest to perform excellently even on small-scale data.

## 2 PROPOSED METHODOLOGY

In this section, we describe the two step process of our proposed methodology, which include extracting features from the image using ResNet50 and then using those features to train a gcForest, in order to perform the task of classification.

### 2.1 FEATURE EXTRACTION USING ResNet50

Since Convolutional neural networks have proved very effective in representation learning, as they extract features through convolutional filters and train the parameters through backpropagation, we use a state-of-the art convolutional neural network, ResNet50[3], pretrained on the ImageNet dataset. The image are resized to 224x224 to fit the input requirements of the ResNet50. The features are obtained by removing the last fully connected layer to get the 2048 dimensional feature vector. These feature vectors were obtained easily without the use of much computational power. Since ResNet50 has the capability to provide effective feature for most images we did not finetune it according to our dataset and instead used the pretrained weights from the ImageNet dataset.

### 2.2 CLASSIFICATION USING gcForest

The features extracted from the ResNet50 is a 2048 dimensional feature vector. These feature vectors are then used to train a gcForest, a novel decision tree ensemble, proposed by Zhi-Hua Zhou *et al.* 2018[4]. We use only the cascade part of the gcForest, as opposed to the multi-grained scanning followed by the cascade part, as the authors in the paper have stated that the multi-grained scanning part is used to capture correlation among spatial data points in an image or sequential data in sequence models. In our model, we have already successfully captured the spatial correlation amongst the data points through the use of the Convolutional neural network ResNet50. Also due to the limited computational resources avaialable we could not model the 2048 dimensional feature vector as temporal data to use a one dimensional sliding window, but we believe that using this will significantly increase the performance of our model. The architecture of the gcForest follows closely the original architecture proposed by the authors of [4]. Each instance is used to generate the 2048 dimensional feature vector. These 2048 dimensional feature vector is then used to train a XGBClassifier, a Random Forest Classifier, a Extra Trees Classifier and a Logistic Regression Classifier. In each layer, each of the 4 classifiers give a 7 dimensional probability distribution vector corresponding to the 7 classes in our problem. These 28 (7 x 4) dimensional vector is then concatenated with the original 2048 features to form a 2076 dimensional feature vector which forms the input for the subsequent layers. The performance of the model is evaluated at each layer, and if no significant improvement in performance is noted, the model does not expand further and returns the last layer with the highest accuracy.

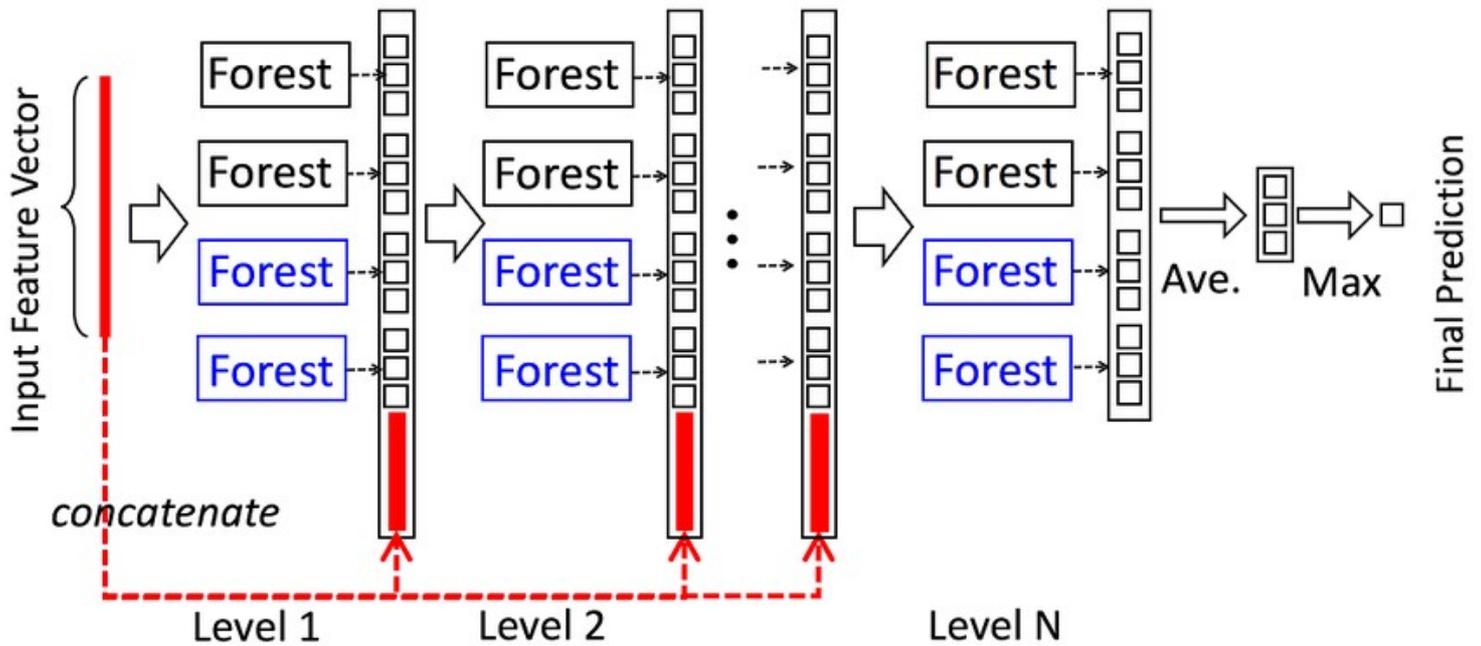

Fig 1. Architecture of gcForest, with the Input Feature Vector having dimension 2048.

The four classifiers shown as Forest here are the Logistic Regression, XGBClassifier, Random Forest Classifier and Extra Trees Classifier.

## 3 RESULTS

We split the dataset into training set and testing set with a split ratio of 0.1. The training accuracy obtained with this split dataset was 97.15%. The test accuracy obtained was 80.04%. This was primarily attributed to the reason that we were using only a 5 fold cross validation. Had we used a 10 fold cross validation the overfitting might have been lesser. But due to limited computational resources we could only use the 5 fold cross validation.

## 4 FURTHER SCOPE OF IMPROVEMENT

We saw that the gcForest did not give very good results as compared to other cases where gcForest was used. One of the main reasons for this could be that the features extracted by the ResNet50 was not a proper representation of the images. We could solve this problem by fine-tuning the layers of ResNet50, where we can train the last few layers of the pretrained network with a lower learning rate. However due to paucity of time we could not explore this region. We leave this as a future scope of improvement.